\documentclass[letterpaper, 10 pt, conference]{ieeeconf}  

\IEEEoverridecommandlockouts                              

\usepackage[utf8]{inputenc}
\usepackage{xcolor}
\usepackage{censor}
\usepackage{afterpage}
\usepackage{graphics} 
\usepackage{epsfig} 
\usepackage{amsmath} 
\usepackage{hyperref}
\usepackage{subcaption}
\usepackage{booktabs}
\usepackage[font=small]{caption}
\usepackage{makecell}
\usepackage{multirow}
\usepackage{placeins}

\usepackage{titlesec}
\let\labelindent\relax
\titlespacing{\subsection}{0pt}{*0.3}{*0.3}
\usepackage{enumitem}
\setlist[itemize]{noitemsep, ,nolistsep,topsep=0pt}
\setlist[enumerate]{noitemsep,nolistsep, topsep=0pt}
\usepackage{tabularx}      
\usepackage{ragged2e}      
\newcolumntype{L}[1]{>{\RaggedRight\arraybackslash}p{#1}} 
\newcolumntype{C}{>{\centering\arraybackslash}X}           
\title{\LARGE \bf
Human-Aware Target Tracking and Navigation: Fusing Kinematic State Estimation with Structural Map Constraints
}\author{
    Sagar Gupta$^{1,2}$, Don Gideon$^{2}$, Seng W. Loke$^{1}$, Kevin Lee$^{1}$ and Bijo Sebastian$^{2}$
    \thanks{$^{1}$Deakin University, Australia}
    \thanks{$^{2}$Indian Institute of Technology Madras, India}
}

\begin{document}

\maketitle
\thispagestyle{empty}
\pagestyle{empty}

\begin{abstract}

Autonomous mobile robots performing person-following tasks often suffer from temporary occlusions and sensor track loss in dynamic environments. This research presents an end-to-end autonomous navigation stack that addresses target occlusion through map-informed spatial reasoning. The proposed system features a multi-modal perception pipeline, fusing deep learning-based visual tracking with 2-dimensional LiDAR point clustering to maintain high-fidelity tracking of a tagged person. A continuous state estimator integrates this perception data with wheel odometry and IMU sensors for stable localization. When the active track is lost due to occlusion, the system activates a map-based recovery framework. Leveraging a predefined topological map, the system executes a graph-based search to propagate the target's last known trajectory along structurally defined walking lanes, adhering to left-hand regional conventions. By generating a discrete set of feasible future trajectories, the robot reasons about potential structural trajectory changes, such as continuing a heading or turning at an intersection. This map-informed prediction is fed directly to the local obstacle avoidance planner, enabling the robot to continue following its target safely and predictably until the person is visually reacquired. Real-world evaluations in dense multi-person environments demonstrate the system's robustness, achieving a 71.4\% target reacquisition success rate during major occlusion events lasting up to 7 seconds.

\end{abstract}

\section{Introduction}

\begin{figure}
    \centering
    \includegraphics[width=0.75\linewidth]{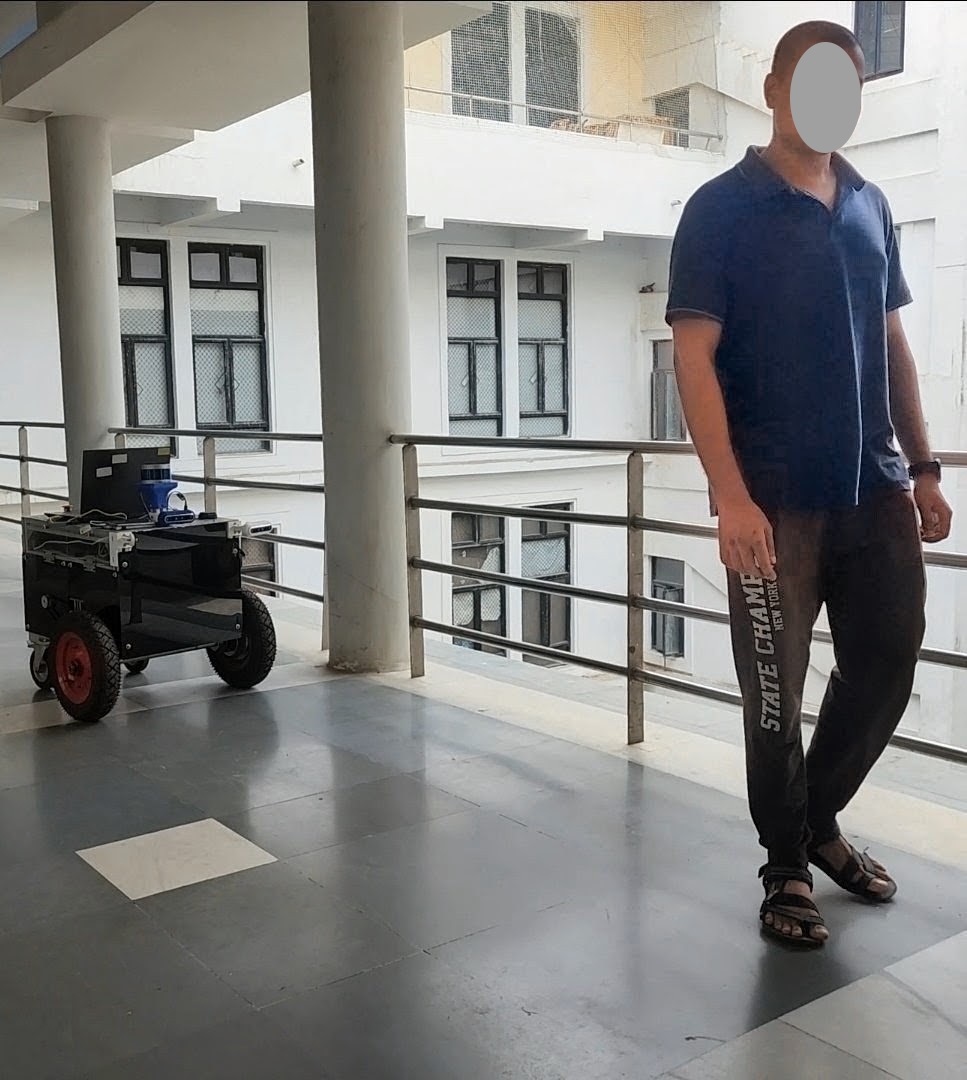}
    \caption{A custom differential drive robot follows a tagged person using vision, LiDAR, and topological map information.}
    \label{fig:g-abs}
\end{figure}

The integration of autonomous mobile robots into human-centric environments requires robust, safe, and socially compliant navigation systems. Among the various tasks assigned to these robots, autonomous person-following remains a critical capability with applications spanning industrial logistics, healthcare assistance, and personal service robotics. As shown in Fig. \ref{fig:g-abs}, our custom differential drive robot leverages multiple modalities to effectively follow a tagged individual. A fundamental requirement of a person-following system requirement of a person-following system is the ability to reliably tag a specific individual and maintain a continuous pursuit. However, in dynamic, real-world indoor environments, robots frequently encounter temporary line-of-sight occlusions—such as the target walking behind obstacles, blending into dense crowds, or turning corners at structural intersections.

\begin{figure*}[t]
    \centering
    \includegraphics[width=0.7\textwidth]{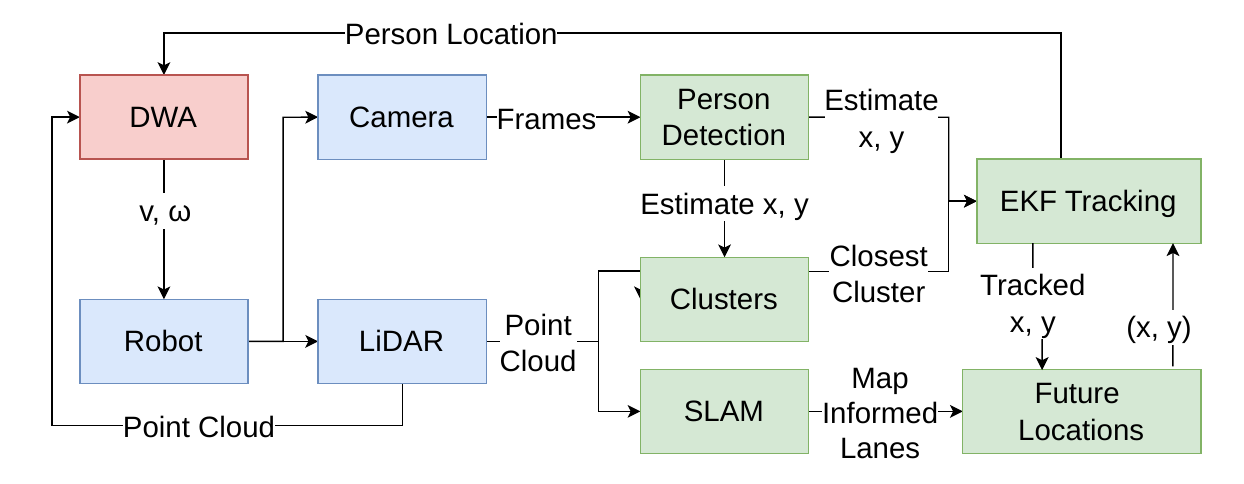}
    \caption{ The proposed system starts with a robot equipped with a camera and a LiDAR. The camera frames are used for person detection using YOLO \cite{yolo} for object detection and DeepSORT \cite{deepsort} for tracking. We have a camera to LiDAR transform, and these tracked boxes are sent as estimates to LiDAR for tagging the closest cluster. The estimate and the closest cluster are fused using an Extended Kalman Filter to continuously track a person. LiDAR's point clouds are also used for SLAM in our indoor environment. We have mapped lanes in which people are likely to walk on the left side, as is convention in left-hand driving countries. In case the EKF track is lost, the map informed lanes are used to estimate future locations \cite{mipp} using the last known position and velocity.}
    \label{fig:sys_overview}
\end{figure*}
Traditional vision-based tracking systems, while adept at identifying targets in clear view, are highly susceptible to failure under occlusion. When a target is obscured, purely visual trackers often lose the subject or suffer from identity switches, especially in densely populated areas where multiple pedestrians share similar visual features. Conversely, relying solely on spatial sensors like LiDAR provides robust structural tracking but lacks the semantic context required to differentiate a specific tagged target from other moving humans. To navigate safely around multiple people while following a specific target, a robot requires multiple sources of verified data. Furthermore, when the sensory track is entirely lost, relying exclusively on short-horizon kinematic predictions (e.g., constant velocity models) often results in unnatural or geometrically impossible trajectory estimates, leading to the ''freezing robot'' problem or complete mission failure.

To address these challenges, this paper proposes an end-to-end autonomous navigation stack designed for reliable person-following under temporal occlusion. Our approach bridges the gap between reactive visual tracking and long-term semantic mapping. We utilize a multi-modal perception pipeline that fuses visual bounding boxes generated by YOLO \cite{yolo} and DeepSORT \cite{deepsort} with 2D LiDAR point clustering, feeding a continuous Extended Kalman Filter (EKF) state estimator. When line-of-sight is broken and the EKF track is lost, our system transitions to a predictive map-based recovery framework. By executing a Breadth-First Search (BFS) on a predefined topological map, the robot propagates the human's last known trajectory along structurally defined walking lanes based on our previous work in \cite{mipp}, actively reasoning about future path choices. This map-informed prediction is fed to the local planner, utilizing DWA \cite{fox1997dynamic} to allow the robot to continue its pursuit proactively until visual reacquisition is achieved.

The primary contributions of this work are summarized as follows:
\begin{enumerate}
    \item The development of a hybrid tracking framework that fuses deep learning-based vision, 2D LiDAR spatial clustering, and topological map data to handle temporary occlusions and track loss.
    \item The introduction of an adaptive occlusion-recovery strategy that transitions from kinematic EKF tracking to structural trajectory prediction via Breadth-First Search (BFS).
    \item Real-world hardware validation on a differential-drive robot, demonstrating the system's ability to reliably tag, track, and pursue a specific individual even in dynamic indoor scenes populated by up to 10 non-target pedestrians.
\end{enumerate}

\section{Related Work}

Maintaining a continuous target track in dynamic environments is a well-established challenge in mobile robotics. The combination of vision-based detection algorithms with appearance-based trackers is widely utilized to maintain target identity over time, providing robust pedestrian tracking by exploiting temporal identity information \cite{mipp, 10.3390/s21030717}. To localize these targets in physical space, heterogeneous sensing modalities—such as vision-derived seed points and LiDAR clustering—are frequently fused \cite{mipp}. The Extended Kalman Filter (EKF) serves as a common baseline for fusing these nonlinear measurements with unicycle-like motion models to track mobile agents \cite{10.1155/2010/482972, 10.1109/iros.2017.8206066, 10.1109/iros.2008.4651188}. The projection of 2D bounding boxes into spatial seeds via a calibrated camera-to-LiDAR transform, followed by point cloud clustering, is a proven methodology for cross-modal data association and target localization in cluttered indoor environments \cite{10.1109/iccsce.2012.6487193, 10.1002/rsa.20443}. Furthermore, recent research highlights the significant potential of fusing alternative modalities to overcome visual blockages, such as integrating auditory features with vision-based detection to reliably identify traffic light states under severe visual occlusion \cite{10802855}.

When sensor measurements are interrupted due to temporary occlusions, state estimation systems typically employ a ''predict-only'' fallback strategy. In this paradigm, the EKF coasts on the last reliable motion model until new observations are available \cite{10.48550/arxiv.1510.06263, 10.1177/1729881418775854}. The design of unicycle-based EKF formulations for 2D mobile robots frequently tracks state variables such as planar position, heading, and linear velocity, inferring the latter two from temporal state propagation. Proper tuning of the measurement and process noise covariance matrices is critical in these systems to balance sensitivity to sensor jitter against responsiveness to sudden motion changes, ensuring stability and consistency in nonlinear estimation \cite{10.1109/cdc.2018.8619726, 10.3390/s17122810}.

While purely kinematic predictions are useful for brief occlusions, they struggle to model complex behaviors over extended periods. Recent literature emphasizes the integration of topological graph information to constrain human-motion predictions \cite{mipp,10.1177/0278364917710541, 10.1002/rnc.3989}. Given prior knowledge of a structured indoor environment, robots can leverage a multi-hypothesis framework to reason about potential structural trajectory changes, such as maintaining a heading or initiating a turn at an intersection \cite{mipp}. Using a predefined topological map to generate a discrete set of feasible future trajectories allows systems to exploit geometric constraints and social norms, such as keeping to the left side of a corridor. Moreover, managing the paths of dynamic agents in shared spaces requires proactive conflict resolution to prevent gridlock; hybrid coordination systems can aggregate decentralized path plans to detect intersection conflicts and issue non-binding stop commands, effectively acting as virtual traffic lights \cite{gupta2025virtual}. By integrating these multi-hypothesis predictions into uncertainty-aware local planners, such as the Dynamic Window Approach (DWA) \cite{mipp, 10.1109/robot.2007.363679, 10.48550/arxiv.2201.01483}, robots can execute proactive, collision-free, and socially compliant navigation maneuvers under partial observability.

\section{System Overview}

\begin{figure*}[!t]
    \centering
    \includegraphics[width=0.9\textwidth]{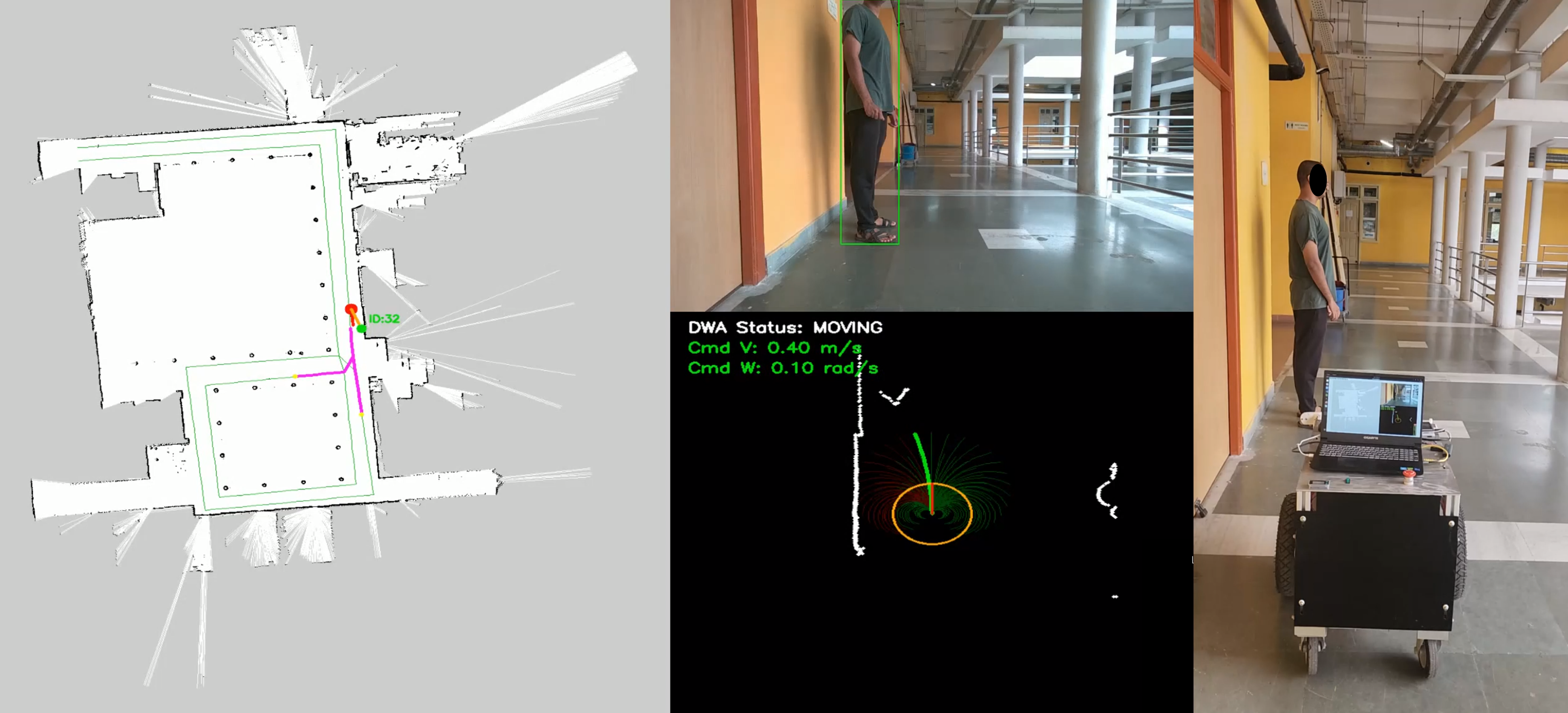}
    \caption{Comprehensive system dashboard during a real-world pursuit scenario. \textbf{Left:} The topological map displays the robot's localized state (red marker), the line-of-sight path to the target (orange line), and purple map-informed predictions of the target's future path adhering to intersection walking conventions. \textbf{Top-Center:} The primary RGB feed indicating the tagged target via a green bounding box. \textbf{Bottom-Center:} The Dynamic Window Approach (DWA) local planner interface, overlaying LiDAR point clouds (white dots) with evaluated velocity arcs. Green trajectories denote traversable paths, red indicates collisions, and the bold green line represents the chosen velocity command. The robot's safety inflation radius is marked by the orange circle, alongside a localized snapshot of the target tracking on the bottom right. \textbf{Right:} A third-person perspective of the platform pursuing the pedestrian through the corridor.}
    \label{fig:full_view_dashboard}
\end{figure*}

The proposed three-tier architecture (Fig. \ref{fig:sys_overview}) fuses high-level visual perception with low-level spatial clustering and continuous state estimation for reliable tracking under occlusion. The software architecture comprises four interconnected modules: a multi-modal perception pipeline for detecting and tracking targets; a robust localization stack fusing sensor data via an Extended Kalman Filter (EKF) and Adaptive Monte Carlo Localization (AMCL); a map-informed occlusion recovery module; and a Dynamic Window Approach (DWA) local planner for collision-free navigation. The supporting hardware suite, detailed in Section IV-A, relies on an RGB-D camera, a 2D LiDAR, an IMU, and wheel encoders. A comprehensive view of the system's operational dashboard, integrating these multi-modal inputs during an active pursuit, is presented in Fig. \ref{fig:full_view_dashboard}.

\subsection{Vision}

The perception module serves as the primary mechanism for identifying and maintaining a lock on the human target. This subsystem extracts 2D bounding boxes from raw RGB frames and associates them across time before projecting them into the LiDAR frame.

\subsubsection{Object Detection}

\begin{figure}
    \centering
    \includegraphics[width=0.7\linewidth]{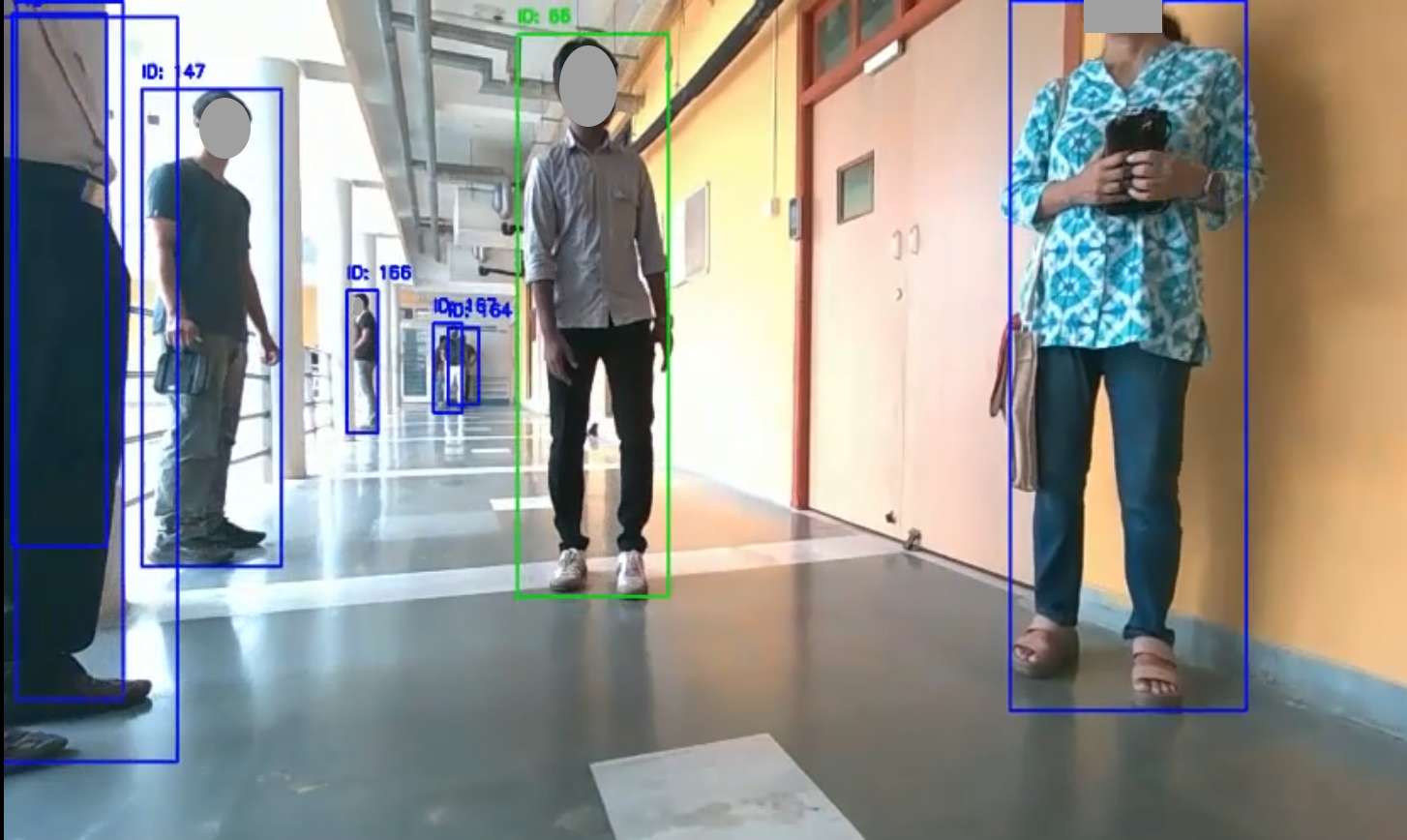}
    \caption{Vision-based tracking of the tagged person, represented in our stack using the green bounding box using the ID:55 provided by the DeepSORT model. The other untagged people are represented using the blue bounding box. This tag is maintained over the course of the experiments, allowing the robot to follow the person.}
    \label{fig:vision}
\end{figure}

The perception pipeline relies on a lightweight YOLOv8 nano model for high-speed, real-time human detection. Executed directly on the primary RGB stream, this model generates precise 2D bounding boxes around pedestrians within the camera's field of view. Prioritizing a lightweight architecture ensures low-latency updates, which is critical for minimizing the reaction time of the robot's low-level controllers. An example of this vision-based tagging in our interface is shown in Fig. \ref{fig:vision}.

\subsubsection{Object Tracking}
\begin{figure}
    \centering
    \includegraphics[width=0.78\linewidth]{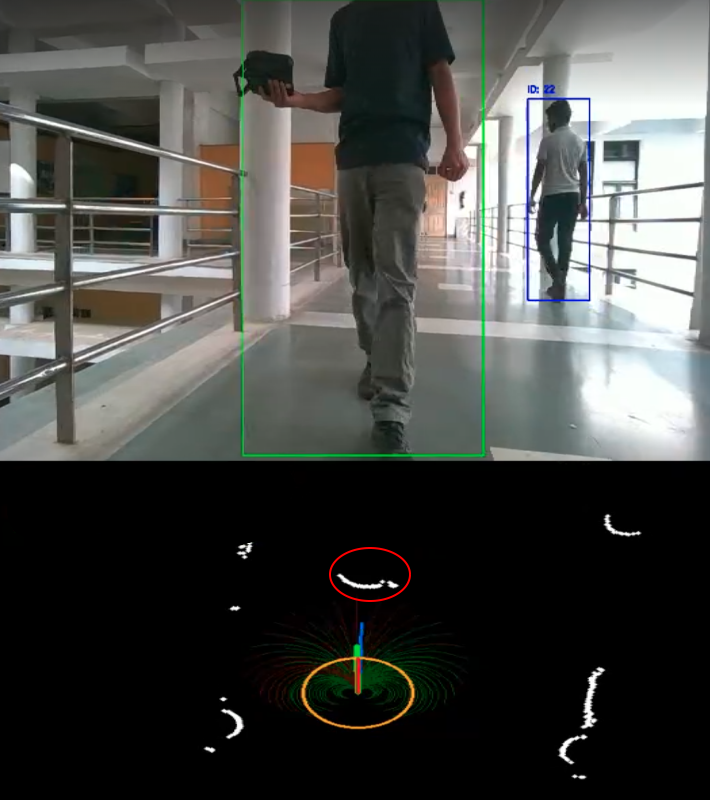}
        \caption{The top view shows the vision feed, as described in Fig. \ref{fig:vision}. The bottom view shows the LiDAR laser scan. The white dots represent the point clouds. The green velocities are traversable, and red ones represent obstacles. The bold green line is the chosen velocity. The orange circle is the robot's inflation radius. The red circle denotes the tracked cluster.}
    \label{fig:lidar}
\end{figure}
The initial person to be tracked is tagged using an openCV dashboard visualizing the various people within the scene using the bounding boxes provided by our object detection model. To overcome the limitations of raw frame-by-frame detection in multi-human environments, the visual subsystem pipelines the YOLOv8 bounding box outputs directly into a DeepSORT tracking algorithm. DeepSORT assigns and maintains consistent, unique identification numbers (IDs) for each detected person across sequential frames by leveraging both spatial Kalman filtering and visual feature extraction. This temporal ID consistency makes the robot highly robust against temporary visual occlusions, such as when pedestrians cross paths in front of the camera or when the target briefly steps behind an obstacle. Operationally, this tracking logic is managed by a centralized perception node, acting as the ''brain'' of the following behavior.

\begin{figure}[htbp]
    \centering
    \includegraphics[width=0.65\linewidth]{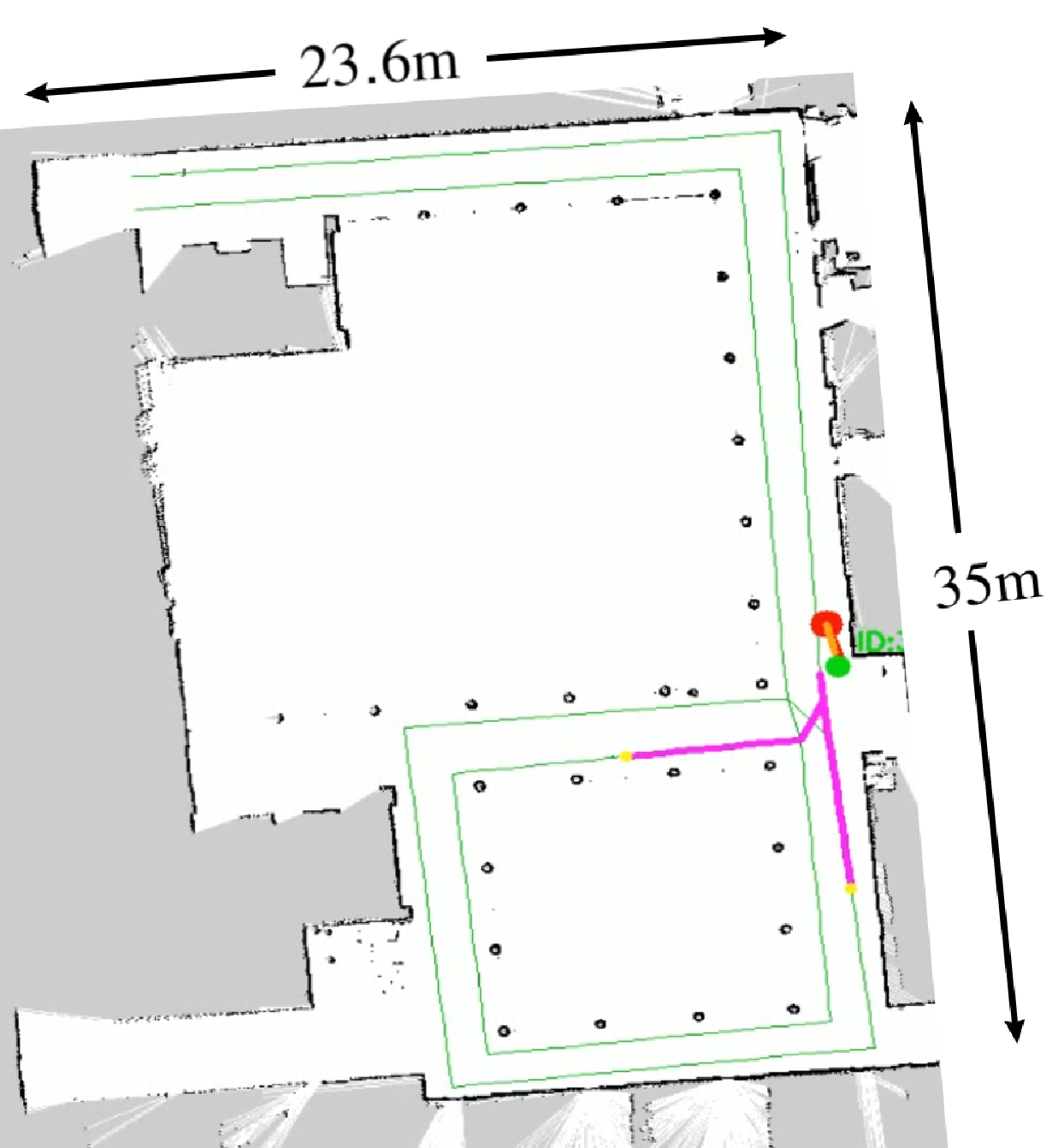}
    \caption{Map-informed track propagation of a tagged person. The green dot represents the tagged person, and the orange line connecting to it is the straight line path from the robot. The purple lanes indicate the future propagations of the target's path based on walking conventions, with lanes diverging through an intersection in the indoor corridors. The system uses these map-informed branches to predict where the person went when line-of-sight is lost.}
    \label{fig:map_informed_bfs}
\end{figure}
\subsection{LiDAR-based tracking}
While the vision subsystem isolates targets in 2D pixel space, local planners require precise 2D spatial coordinates. We bridge this gap using homography and LiDAR spatial clustering. A pre-calibrated homography matrix first projects the target's 2D bounding box onto the ground plane. Because homography is sensitive to camera pitch, this projection merely acts as a spatial seed to interrogate a downsampled 2D planar laser scan from the LiDAR.

\subsubsection{Clustering and Search}
The system relies on a Velodyne VLP-16 LiDAR, mounted and mapped within the robot's static transform (TF) tree. To ensure real-time responsiveness and optimize computational efficiency, the raw 3D point cloud generated by the Velodyne VLP-16 LiDAR is first downsampled into a 2D planar laser scan. The vision-based perception node searches the high-resolution point-cloud within a strict $0.5$-meter radius of the initial homography seed. By clustering the LiDAR points within this localized search window, the system successfully filters out background noise and isolates the physical, geometric body of the target. This fusion produces an accurate center-point coordinate for the person. Finally, to bridge the gap between discrete LiDAR scans and continuous robot motion, each clustered person is tracked using an internal 4-state Extended Kalman Filter estimating $(x, y, \theta, v)$. If the LiDAR momentarily loses the physical cluster, this target-level EKF allows the system to coast and predict the human's immediate trajectory based on their last known velocity, maintaining a stable setpoint for the downstream path generation. By clustering the LiDAR points that fall within the localized search window passed down by the vision-based estimates, the system rejects environmental background noise and resolves visual depth ambiguities, yielding an accurate, discrete $(x, y)$ coordinate representing the target's true physical center in the global frame. This clustered target and the evaluated navigational velocity arcs are visualized in Fig. \ref{fig:lidar}.

\subsubsection{Extended Kalman Filter}

To bridge the gap between discrete, frame-by-frame LiDAR clusters and the continuous motion required for smooth robotic navigation, each detected person is tracked using a custom Extended Kalman Filter (EKF). This filter maintains smooth trajectory tracking and estimates human motion during occlusions when sensors lose targets. The EKF tracks a 4-dimensional state vector for each person, defined as $\mathbf{X} = [x, y, \theta, v]^T$, where $x$ and $y$ represent the global planar coordinates, $\theta$ denotes the heading in radians, and $v$ is the linear velocity in meters per second. When a new person is detected, the estimate covariance matrix is initialized with moderate confidence, $\mathbf{P} = \mathbf{I}_{4 \times 4} \times 0.1$. 

The filter operates through a standard two-step prediction and update mechanism. During the prediction step, the target's state is propagated forward in time using a non-linear unicycle kinematic model:
\begin{align*}
    x_{new} &= x + v \cos(\theta) \Delta t \\
    y_{new} &= y + v \sin(\theta) \Delta t
\end{align*}
To accommodate the unpredictability of human motion, the process noise covariance matrix is tuned as $\mathbf{Q} = \text{diag}([0.05, 0.05, 0.1, 0.1])$. Lower uncertainty ($0.05$) is assigned to physical spatial changes, while higher uncertainty ($0.1$) is applied to heading ($\theta$) and velocity ($v$). This tuning reflects a human's physical capability to stop, accelerate, or pivot abruptly, prompting the filter to trust sensor measurements over the strict kinematic model for speed and direction changes. During the update step, the observation matrix $\mathbf{H}$ isolates only the $x$ and $y$ spatial coordinates, as the LiDAR clustering algorithm does not directly measure velocity or heading. The filter implicitly deduces $v$ and $\theta$ from the temporal changes in position. The measurement noise covariance matrix, $\mathbf{R} = \text{diag}([0.1, 0.1])$, is set slightly higher than the spatial process noise to smoothly average out the inherent jitter from the raw LiDAR measurements, preventing the tracked center from snapping erratically to minor cluster fluctuations. To handle occlusions, the EKF skips LiDAR measurement updates and coasts targets forward using their last known velocity, resetting to the camera's visual projection with inflated covariance if predictions drift over 1.0 meter. If LiDAR updates fail to re-establish within 5.0 consecutive seconds, the tracker is deemed lost and pruned to conserve memory.

\begin{figure}[!t]
    \centering
    \includegraphics[width=0.65\linewidth]{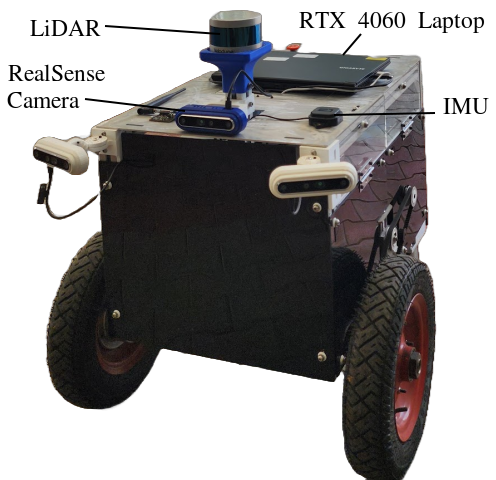}
    \caption{The custom differential-drive mobile robot utilized for real-world hardware validation, equipped with an Intel RealSense RGB-D camera and a Velodyne VLP-16 LiDAR.}
    \label{fig:hardware}
\end{figure}

\subsection{Map-Informed Spatial Reasoning}

When the primary perception suite (vision and LiDAR) loses the tracked target—often due to the person turning a corner or stepping behind an obstacle—relying solely on the EKF's short-horizon kinematic tracking is insufficient. Standard constant velocity models often generate implausible predictions in structured indoor environments, such as passing through walls or failing to anticipate turns. To ensure continuous tracking, the system transitions to a map-informed spatial reasoning framework. Given prior knowledge of a structured indoor environment and a person's position, a robot can leverage a multi-hypothesis framework to reason about potential structural trajectory changes, such as maintaining a heading or initiating a turn at an intersection~\cite{mipp}. The system integrates map information using a topological graph, which abstracts the environment into key nodes (e.g., intersections, corners) and traversable directed edges~\cite{mipp}. Upon losing the sensory track, a breadth-first search (BFS) is employed to continue the prediction along the map, up to a predefined maximum path length~\cite{mipp}. 

Using this predefined topological map, the system generates a discrete set of feasible future trajectories for humans around the robot. This graph-based propagation respects geometric constraints and social norms, such as the regional convention of staying on the left side of a corridor. The BFS explicitly reasons about the nature of the occlusion itself. If the topological graph branches at an intersection, the system evaluates the diverging paths. A path that continues straight may be deprioritized or ignored if it remains within the robot's clear line of sight, whereas a turning path is actively explored and propagated because it presents a much higher probability of having caused the occlusion.

\section{Experiments}

\subsection{Hardware Setup}
Experiments were conducted on a custom differential-drive mobile robot, pictured in Fig. \ref{fig:hardware}. The platform features a $0.65\text{ m}$ wheelbase and utilizes a conservative $0.45\text{ m}$ circular footprint for safety inflation layers. Locomotion is managed via an \texttt{Arduino}-controlled PI-feedforward loop capped at $v_{\max} = 0.4\text{ m/s}$ and $\omega_{\max} = 1.5\text{ rad/s}$. The perception suite comprises an Intel RealSense camera for YOLOv8-based human tracking, a Velodyne VLP-16 LiDAR downsampled to a 2D planar scan, and an Xsens MTi IMU.

\subsection{Indoor Environment}
The real-world evaluations were conducted within a structured indoor environment characterized by long, orthogonal corridors and three-way intersections. As previously depicted in the topological mapping (see Fig. \ref{fig:map_informed_bfs}), this environment enforces specific spatial constraints and behavioral norms on its occupants, primarily the convention of walking on the left side of the hallway\cite{mipp}. These structural constraints serve as the foundation for the system's map-informed target prediction during occlusion events. \footnote{Video demonstrations of the real-world trials are available at: \url{https://youtu.be/vMKtNnshklE}}

\subsection{Experimental Scenarios}

To rigorously test the limits of the tracking and recovery pipeline, the system was evaluated across 5 distinct trials with varying pedestrian densities. Across all trials, the robot operated for a total duration of 20 minutes. For image processing and navigation, we used a laptop equipped with an RTX 4060 with 8GB VRAM. The laptop ran YOLOv8n on RealSense feeds downsampled to 1280 × 720 resolution, producing an output of 16.2 frames per second (FPS) averaged over the runs. The DeepSORT model processed 100\% of the bounding boxes provided by the object detection model.

\textbf{Single Target with Erratic Maneuvers:} In the foundational scenario (1 person), the robot was tasked with tracking a single individual navigating the corridor network. To stress-test the continuous EKF tracking and spatial reasoning modules, the target deliberately executed erratic movements, including abrupt turns around corners, sudden stops, and lateral side-to-side walking. 

\begin{figure}[t]
  \centering
  \begin{subfigure}[t]{0.48\linewidth}
    \centering
    \includegraphics[width=\linewidth]{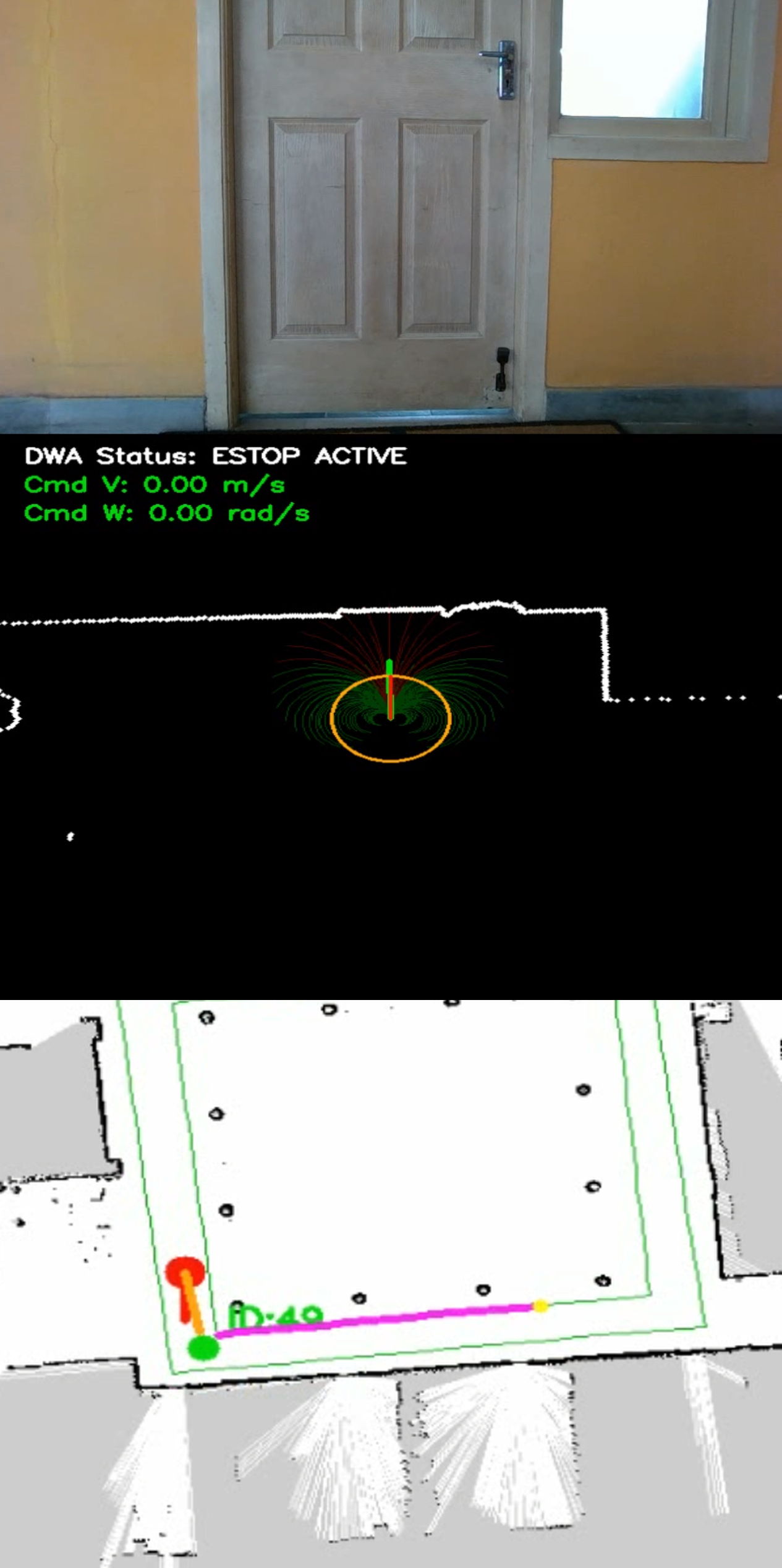}
    \caption{The robot loses track of the person in camera and LiDAR. In the bottom map view, we see the last seen track of the person (green dot), and their future path (purple path).}
    \label{fig:occ1}
  \end{subfigure}\hfill
  \begin{subfigure}[t]{0.45\linewidth}
    \centering
    \includegraphics[width=\linewidth]{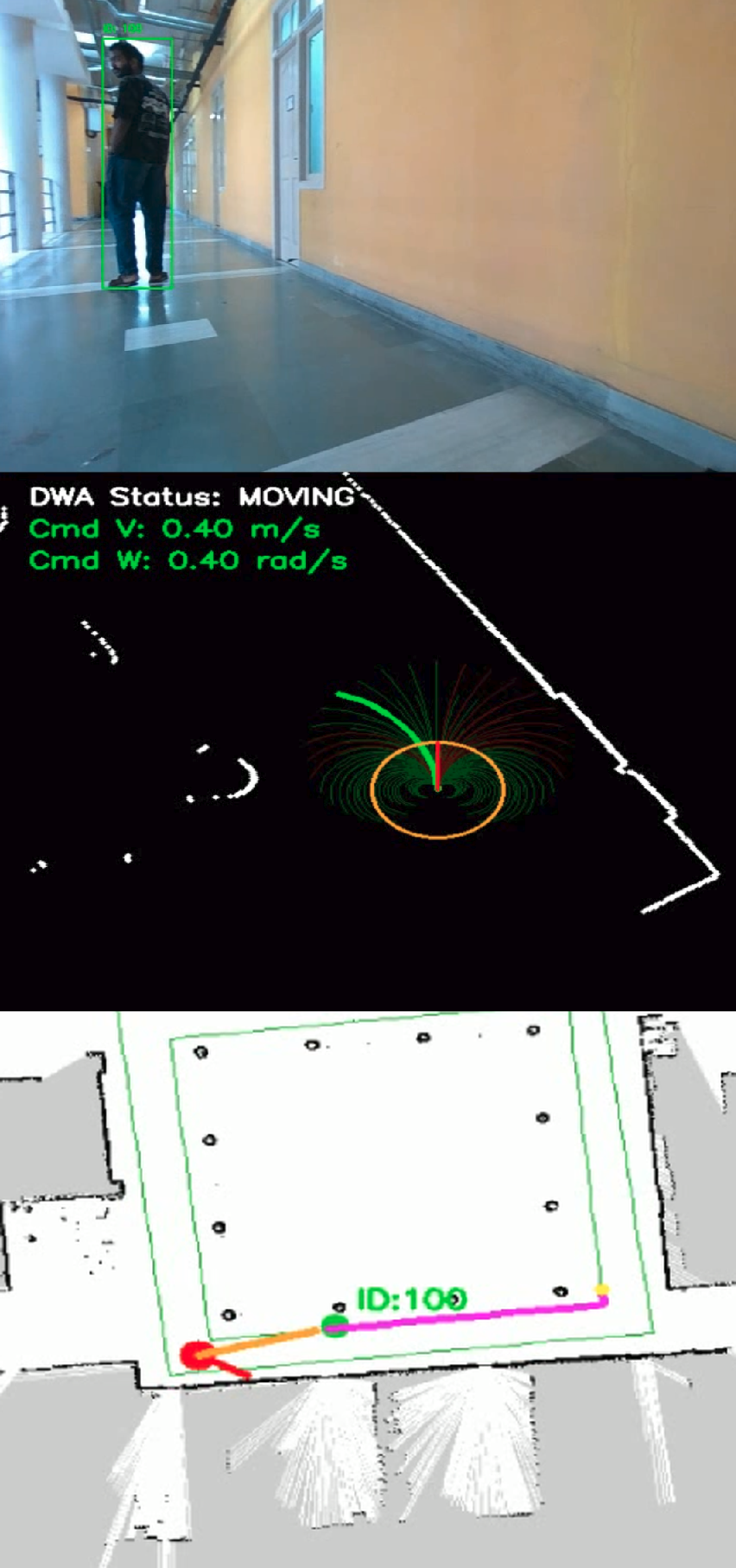}
    \caption{The robot decides to turn left to find the person, and identifies in both the camera and LiDAR frame, but with a different DeepSORT ID, as it is a re-association.}
    \label{fig:occ2}
  \end{subfigure}
  \caption{Example occlusion events from real-world trials where the robot used map-based propagation to predict where the person might be. On the top, we see the first person view of the camera, with the LiDAR clusters shown in the middle view. The bottom view shows the map.}
  \label{fig:occlusions}
\end{figure}

\textbf{Dense Multi-Person Environments:} To evaluate the system's robustness against complex, dynamic occlusions, the platform was deployed across the remaining four scenes featuring multiple non-target pedestrians (densities of 4, 6, 7, and 11 people). In these dense scenarios, the system encountered 7 major occlusion events, with visual blockage durations ranging from a minimum of 1 second to a maximum of 7 seconds. Examples of these challenging scenarios are depicted in Fig. \ref{fig:occlusions}, which illustrates the initial loss of sensory track (Fig. \ref{fig:occ1}) and the successful map-based reacquisition as the robot navigates to intercept the predicted path (Fig. \ref{fig:occ2}). The map-informed recovery framework achieved a 71.4\% reacquisition success rate, with the time-to-reacquire taking up to 10 seconds per event. While the system demonstrated high resilience to visual interference by leveraging structural predictions, the challenging environments did induce some tracking errors, resulting in 5 ID-switches and 2 complete re-tag failures over the course of the multi-person scenes.

\section{Conclusion}
This paper presented an end-to-end autonomous navigation architecture designed to solve the persistent challenge of target occlusion in robotic person-following. By bridging the gap between reactive sensor tracking and long-term spatial awareness, the proposed system fuses deep learning-based visual inference with LiDAR point clustering and continuous EKF state estimation. To overcome the limitations of purely kinematic trajectory predictions during line-of-sight loss, we introduced a map-informed recovery framework. Leveraging a predefined topological map, the system dynamically reasons about structural path choices, propagating the target's trajectory along expected walking lanes to proactively anticipate movements outside its immediate sensory field of view. Real-world experiments validated our mobile robot's ability to track targets through erratic maneuvers and dense crowds. Future work will integrate dynamic crowd-flow prediction to enhance adaptability in unstructured environments.

\bibliographystyle{IEEEtran}
\bibliography{ref.bib}

\end{document}